# Explainable and Human-Grounded AI for Decision Support Systems: The Theory of Epistemic Quasi-Partnerships

John Dorsch[1] & Maximilian Moll[2]

1. Faculty of Philosophy, Philosophy of Science and the Study of Religion, Ludwig Maximilian University Munich
2. Faculty of Computer Science, University of the Bundeswehr Munich

Abstract

In the context of AI decision support systems (AI-DSS), we argue that meeting the demands of ethical and explainable AI (XAI) is about developing AI-DSS to provide human decision-makers with three types of human-grounded explanations: reasons, counterfactuals, and confidence, an approach we refer to as the RCC approach. We begin by reviewing current empirical XAI literature that investigates the relationship between various methods for generating model explanations (e.g., LIME, SHAP, Anchors), the perceived trustworthiness of the model, and end-user accuracy. We demonstrate how current theories about what constitutes good human-grounded reasons either do not adequately explain this evidence or do not offer sound ethical advice for development. Thus, we offer a novel theory of human-machine interaction: the theory of epistemic quasi-partnerships (EQP). Finally, we motivate adopting EQP and demonstrate how it explains the empirical evidence, offers sound ethical advice, and entails adopting the RCC approach.

# 1. Introduction

In an era when AI decision support systems (AI-DSS) play an ever-increasing role in high-stakes environments, the need for ethical decision-making support technology takes center stage. This demand is made clear by the integration of AI-DSS into child welfare services for assisting the decision to investigate the possibility of child maltreatment (Kawakami et al. 2022). While controversy exists around whether machines can be genuinely trustworthy, since, for example, machines cannot possibly be our peers (Bryson 2018), orthogonal to these concerns is the *perceived trust* in the system: the trustworthiness that deployers (e.g., end-users) attribute to the system, whether the system is trustworthy or not. Perceived trust is worthwhile to increase as it can serve to mitigate algorithm aversion, wherein deployers reject the model's contributions to the decision space while knowing contributions constitute crucial information which ought to bear on the ultimate decision (e.g., Dietvorst et al. 2015).[1]

As the demand for trustworthy AI gains momentum, research has turned to the field of explainable AI (XAI) (e.g., Miller et al. 2022).[2] This subdiscipline has the mission of designing tools like AI-DSS to be perceived as trustworthy by ensuring the model's decision-making processes are made transparent. This pursuit aligns with the EU AI Act 2024, which mandates that, "High-risk AI systems shall be designed and developed in such a way to ensure that their operation is sufficiently transparent to deployers. This includes providing adequate information to enable deployers to understand the system's capabilities and limitations" (European Union 2024, §13).

However, the Act leaves open the question of what it means to be *sufficiently transparent* to deployers, who typically lack fluency with machine learning methods required to utilize explanations fully, and XAI struggles to offer an accurate, ethical answer. This is because XAI is threatened by the "inmates running the asylum problem" (Miller et al. 2017), wherein explanations are designed for developers rather than for typical deployers.[3] Moreover, standardized evaluation metrics for explanations are lacking (Nauta et al. 2023), meaning XAI researchers must often rely upon intuition to support what constitutes good explanations (Miller 2019). For this reason, Doshi-Velez and Kim (2018) introduced the notion of *human-grounded explanations* to refer to explanations well-suited for deployers, arguing that intuitions ought to be validated by empirical studies with simplified models.

---

[1] Dorsch and Deroy (2024) have provided an argument against developing trustworthy AI, demonstrating it is unnecessary for mitigating the vulnerability that arises when AI-DSS are deployed in medium to high-stakes scenarios. That said, their argument concerns *moral trustworthiness*, which they explain is distinct from *epistemic trustworthiness*. As we will detail in Section 2, our argument concerns the usefulness of *epistemically trustworthy* AI, which means our argument is not in conflict with the views of these authors. That said, a complete defense of our proposal as securing a genuine kind of trustworthiness would go beyond this discussion's scope, so we will leave this for future research.

[2] To be clear, by 'AI' we mean exclusively machine learning systems, including both supervised and unsupervised learning models, neural networks, and other algorithmic approaches used within the realm of machine learning. For the purpose of this discussion, we exclude reinforcement learning because the subset of the literature discussed below does not involve models developed using reinforcement learning techniques.

Thus, a growing consensus is emerging that explainability be distinguished from interpretability (see, e.g., Amazon Web Services 2024). Interpretability is about, "understanding how the model makes decisions, based on a holistic view of its features and each of the learned components such as weights, other parameters, and structures" (e.g., Molnar 2022, §3.32). Explainability goes beyond interpretability by generating deployer-understandable insights into how the model arrives at specific outcomes. To that end, we suggest re-envisioning this field as composed of two subsets, XAI (explainable AI) and IAI (interpretable AI), and a superset *interpretable and explainable AI* (IX-AI). We propose recasting the task of developing XAI, as it concerns the development of AI-DSS, to be about determining how best to craft interpretations to serve as useful explanations (or any additional explanations which are not, strictly speaking, interpretations) and how best to communicate these explanations to deployers, relative to their level of expertise in machine learning. That said, we will refer to *human-grounded explanations* to ensure focus remains on the deployer end of XAI research.[4]

In what follows, we review a subset of the empirical literature in XAI research on human-grounded explanations and present key findings that serve as explanatory targets for related theories, specifically the subset which handles the income prediction task or has a significant bearing on those results (**Section 2**).[5] Thereafter, we canvass three prominent theories that purport to explain what constitutes good human-grounded explanations and argue that they do not adequately explain the empirical evidence or do not impart sound ethical advice for development (**Section 3**). Consequently, we present a novel theory for what constitutes good human-grounded explanations, the *theory of*

---

[3] Strictly, speaking a deployer is any natural or legal person who deploys an AI system in a professional setting (EU AI Act 2024). When we refer to deployers, we mean typical deployers: human persons with domain knowledge, albeit little to no knowledge of machine learning techniques.

[4] Of course, machine learning experts are also humans. Doshi-Velez and Kim's reference to *human*-grounded reasons is meant to emphasize aligning explanations with ordinary epistemic practices (see Section 3 for more).

[5] Given its essentially nonexistent definition, the field of AI subsumes many different algorithmic approaches. In the field of XAI the primary focus is on Machine Learning algorithms, since their behavior cannot be easily understood. However, even within Machine Learning, the type of algorithm can vary greatly (for example Supervised Learning compared to Reinforcement Learning) as well as the type of data can vary (numerical data vs time series). Each of these variations naturally requires different approaches to explanations, which introduces many more hard-to-control variables. As such, we focus here on Supervised Learning approaches that have been evaluated on the same problem. Extending our results to further settings is left for future work.

*epistemic quasi-partnerships* (EQP) (**Section 4**). Though a full defense of EQP will go beyond the scope of the present discussion, we motivate adopting EQP as an ethically sound theory and argue for it by demonstrating how it explains the empirical evidence. To that end, developing ethical and explainable AI-DSS is about designing epistemically trustworthy quasi-partners[6] that provide deployers with human-grounded explanations facilitating common epistemic practices, which we call *the reasons, counterfactuals, and confidence approach* (RCC). Thus, to fill this research lacuna and make sense of the empirical evidence, we defend a conceptual argument for the RCC approach and offer EQP as a theory of human-machine interaction.

## 2. Review of Empirical Studies on Human-Grounded Explanations in AI-DSS

### 2.1 Trust Calibration and the Switch Percentage Paradigm

The effective use of AI-DSS depends on deployers perceiving them as trustworthy (for a comprehensive review, see Zieglmeier & Lehene 2021). As is commonly understood in XAI literature, perceived trustworthiness in the system refers to one of two observable measures. On the one hand, *subjective measures* refer to deployers reports that they perceive the system as trustworthy (the precise question has multiple variants). On the other hand, *objective measures* track certain trust-based behaviors during the human agent's use of the system.[7] One trust-based behavior particularly crucial for measuring algorithm aversion is the act of changing one's own prediction to the system's prediction, measured by the switch percentage (e.g., Ribeiro et al. 2016).

Thus, the system's perceived trustworthiness can be objectively measured by deploying the switch percentage paradigm, which has the goal of determining the enabling factors for optimizing trust calibration (also known as 'appropriate reliance') (Lee & See 2004). Trust calibration represents the degree to which the human agent follows the system's recommendation when accurate (which encompasses cases where the human defers to the system's recommendation), combined with the degree to which the human agent deviates from the system's recommendation when inaccurate. Trust calibration is thus a delicate balance between over-trusting and distrusting the system, and the switch percentage offers insights into how to optimize trust calibration, namely by designing the system to incorporate the enabling factors for switching, while, of course, maintaining system accuracy independently of its perceived trustworthiness.

---

[6] Much will hinge on what exactly it means to say that the partnership is only a quasi one, and this will be discussed in some detail below (**Section 4**). For now, it helps to point out that quasi means *as though*, so that these partnerships are not genuine partnerships, but useful fictions.
[7] This distinction parallels the well-known difference between stated and revealed preferences in economic theory (Samuelson 1938).

**2.2 Explanation Techniques in AI-DSS: Feature Graphs and Model Confidence**

The income prediction task (IPT) is deployed to investigate experimentally the factors that modulate trust calibration and mitigate algorithm aversion. As such, IPT represents an influential paradigm in XAI research to assess and test theories about human-grounded explanations. In this task, participants predict an individual's income level based on various attributes, such as education, occupation, age, etc. To set up the task, researchers train a machine learning model on a relevant dataset (usually a recent US consensus). While models tend to be diverse (decision trees, random forests, neural networks, etc.), each model has been designed with the goal of predicting income levels, achieving a relevant threshold of accuracy for the purpose of the experimental design.

After training, various XAI techniques are employed to generate explanations for the model's predictions, predominantly LIME (Local Interpretable Model-agnostic Explanations) and SHAP (Shapely Additive exPlanations) (e.g., Ribeiro et al. 2016). Both methods elucidate contributions of individual features to the model's prediction. Shapley values, rooted in cooperative game theory, provide a fair distribution of feature importance by considering the average contribution of each feature over all possible subsets of remaining features, though these explanations are computationally intensive. LIME, on the other hand, approximates the model locally using a simpler surrogate model, offering faster and less complex computations but potentially less comprehensive feature interaction insights. Despite their differing approaches, both methods can be similarly used to represent feature importance visually, often bar charts to illustrate the impact of each feature on specific predictions. For this reason, we shall refer to this family of explanation as 'feature graphs'. Finally, researchers set up experimental conditions to observe behavioral results contrasted across different explanation conditions (e.g., presenting or not presenting this explanation or that explanation).

Before analyzing the relevant studies, an important caveat should be issued regarding what these studies measure. Objective measures, such as the switch percentage, do not measure participants' attitudes about the trustworthiness of the system, but rather the degree to which participants switch their original prediction to the system's prediction. In other words, the switch percentage measures the frequency with which participants *defer* to AI-DSS. This can be understood as an act of trust, so long as deferring is constituted by some vulnerability on the part of the deployer (in these studies, vulnerability is often represented by monetary wagers on the accuracy of the prediction).[8] But it would be inaccurate to conceive of this trust as a moral kind. This is because the decision to defer is not justified by appealing to the system's possession of moral rationality, a capacity to act on behalf of moral reasons (for an extended discussion of this, see Dorsch & Deroy 2024). Rather, this trust is *epistemic*, since the decision to defer is justified by appealing to the system's capacity to determine statistical regularities that bear on the relevant facts and inform the ultimate decision (i.e., the prediction).

## 2.3 Case Studies and Comparative Analysis of Explanation Methods

### 2.3.1 The Role of Model Confidence in Trust Calibration

Let us discuss the results of two studies that deployed the switch percentage paradigm to measure trust calibration in IPT, wherein participants were novices in machine learning, akin to typical deployers. First, Zhang and colleagues (2020) found that presenting model confidence in a propositional format facilitated trust calibration over displaying feature graphs. To be clear, propositional format means the explanation resembled a declarative sentence (e.g., "The model is 90% confident that this person earns over $50,000."), meaning the explanation was not unlike the reasons humans provide, which is constituted by truth conditions.[9] Model confidence, on the other hand, refers to a reliability score, specifically the probability assigned by the model to a particular prediction, indicating how certain the model is about the accuracy of its prediction.[10]

Results of this first study raise questions about why model confidence in propositional format is so effective, and why confidence is more effective in modulating epistemic trust than feature graphs. One initial thought is that feature graphs are difficult for novices to understand, but seeing as participants in this study were trained on how to read these graphs, the issue is likely more nuanced than this, requiring a gloss on what it exactly it means to say feature graphs are "difficult to understand". Indeed, we suggest the gloss that these graphs are *epistemically opaque*, that is, they do not transparently convey the *relevant epistemic features behind the model's decision*, leading to difficulties in decision-making and less effective trust calibration (more on this below).

---

[8] Another way of thinking about what is being measured is the difference between trust in the system and trust in oneself. In other words, it measures whether the individual places more trust in the system than in themselves, even if both levels of trust are low. Our thanks to Kristian Barman for pointing this out.
[9] The model's confidence functions as an explanation in the sense that it explains (in part) why it made the prediction it did, because it was confident. Notice that this is conform with ordinary epistemic practices, wherein one explains one's decision by appealing to one's confidence. For example, a doctor might explain their choice of treatment by stating they were confident in the diagnosis based on the patient's symptoms and test results.
[10] Model confidence can be computed in various ways, such as probabilistic confidence scores (local), recall, precision, F1 scores, and area under the curve (global), with probabilistic scores often being deployed in this literature for how they indicate certainty in individual predictions; that said, the present argument is neutral with respect to the best reliability score for human-grounded reasons.

**2.3.2 Feature Graphs and their Epistemic Opacity**

Somewhat surprisingly, Zhang and colleagues found that neither condition (model confidence nor feature graph) led to significant increases in participant accuracy. This outcome suggests at least two possible reasons, each offering important lessons for XAI research. As the researchers explain, both humans and the model had similar error boundaries, meaning conditions where the model had low confidence were also challenging for humans. This indicates that it makes little sense to deploy AI-DSS if their capacity to contribute to the ultimate decision does not complement the human's; nor does it make sense to deploy AI-DSS if humans do not have unique knowledge that would complement the AI-DSS's contribution. Thus, the aim of deploying AI-DSS ought to be about building robust *human-AI quasi-partnerships* of an epistemic kind, wherein the strengths and expertise of both humans and AI-DSS are leveraged to complement each other's weaknesses.

Finally, it is challenging for deployers to infer model confidence from feature graphs. To this point, Zhang and colleagues write, "In theory, prediction confidence could be inferred by summing the positive and negative contributions of all attributes. If the sum is close to zero, then the prediction is not made with confidence" (p. 9). Such an inference demands a technical understanding of how confidence can be calculated, so deployers are unlikely to employ it. For this reason, feature graphs are epistemically opaque: they require technical understanding to determine *reasons* that would support relevant beliefs in decision-making, such as the belief that the model's prediction is correct.[11] Consequently, the evidence shows that feature graphs, albeit visually instructive, remain nonetheless unintuitive for non-experts. If our goal is to comply with the EU AI Act and make the operation of AI-DSS transparent to deployers, we ought to be skeptical of employing feature graphs for this purpose, perhaps better suited to IAI than XAI.

---

[11] The notion that confidence can serve as a reason to believe a proposition aligns with the frameworks articulated by Goldman (1991) and Hardwig (1985) within the domain of social epistemology. In this context, the model is treated as an expert authority, and, under normal circumstances, an expert's confidence in proposition $P$ provides a justifiable reason to believe that $P$.

**2.3.3 Counterfactual Explanations and Comparative Insights**

The second study deeply relevant to our discussion was conducted by Le and colleagues (2022), who explored the role of counterfactual descriptions when coupled with model confidence in modulating trust calibration. This study contrasted two types of counterfactual descriptions, text-based and graphical, contrasted with the control condition which had no explanation whatsoever. These descriptions were generated through a loss function that balanced the distance between original and counterfactual inputs while achieving the desired confidence change. For example, Le and colleagues showed participants that changing "Marital Status" from "Married" to "Divorced" reduced the confidence score significantly.

These researchers found that displaying both types of counterfactual descriptions improved accuracy and increased the switch percentage over the control condition. Meanwhile, the graphical description increased trust calibration and performance slightly more than the textual description. Le and colleagues suggest that this is because the graphical description is more intuitive and easier to understand (i.e., epistemically transparent). So, the lesson is that counterfactual descriptions are effective explanations, and counterfactual explanations are best communicated *graphically,* if possible.

In addition to the effectiveness of counterfactual information, this finding also demonstrates that feature graphs are not ineffective because they are graphs. Indeed, combining the results of both studies yields the proposal that the issue with feature graphs is that they are epistemically opaque because they fail to facilitate *epistemic discernment*, the capacity to determine reasons for the relevant beliefs important for decision-making. Consequently, these studies raise questions about why model confidence (especially in propositional format) and counterfactuals are so effective in facilitating epistemic discernment and thereby increasing epistemic trust, and XAI theories are challenged to explain this (see **Section 3**).

**2.3.4 Comparing Anchor Explanations and Feature Graphs**

Other studies emphasize comparing explanations in terms of whether they increase participant accuracy, rather than epistemic trust. Ribeiro and colleagues (2018) used IPT (among other tasks) to demonstrate that Anchor explanations outperformed LIMEs in increasing participant accuracy as well as their reported confidence in relying on the system's prediction, which held true across various conditions (e.g., different datasets (tabular data, text, images, etc.) and models (logistic regression, neural networks, etc.)). Anchor explanations refer to sets of specific, contextually relevant features that influence the model's prediction, generated after the model has already been trained. For example, in a model predicting whether a text is positive or negative, an Anchor explanation might highlight words like "excellent", "fantastic", and "amazing" as features that lead to a positive sentiment prediction. This model then produces high-precision, if-then rules that specify conditions under which a prediction holds true.

This gloss we wish to place on these results is that Anchor explanations are more effective than feature graphs because they facilitate epistemic discernment. Specifically, if-then rules encourage deployers to apply the fundamental rule of logic, modus ponens (logical rules are truth-preserving, and so function as epistemic reasons), and, as such, Anchor explanations resemble the material conditional of natural deduction, which reflects the processes that humans employ whenever reasoning logically (Gentzen 1964). Thus, Anchor explanations appear epistemically transparent to deployers, meaning it is clear how these explanations can and ought to be employed in their natural way of processing information and decision-making. As a result, we can specify the goal of making the operation of models sufficiently transparent to deployers as amounting to offering explanations that can be fluently employed in ordinary epistemic practices (see **Section 4**).

Furthermore, Weerts and colleagues (2019) found that feature graphs were employed differently depending on how intuitive they were perceived to be. If very intuitive, then feature graphs were handled as support for model predictions. But if feature graphs were unintuitive, they were treated as evidence that model confidence should be lower. This finding reinforces our proposal that feature graphs are epistemically opaque, since their efficacy in modulating epistemic trust appears to depend on the degree to which they are perceived as intuitive, not accurate. Somewhat worryingly, this result might suggest that feature graphs introduce a form of bias, wherein epistemic trust in the model is unduly influenced by how intuitive deployers find the graphs, rather than by an objective assessment of the model's reliability.

To determine how well deployers understand the functionality of the model, Hase and Bansal (2020) investigated simulatability, which is the degree to which participants accurately predict the behavior of the model when provided with novel or perturbed inputs. In stark contrast to the above studies, Hase and Bansal found that LIME-based feature graphs with tabular data outperformed other kinds of explanations in facilitating participants' simulatability, including Anchors, decision boundaries, and composites of these kinds. The one exception to this finding is that prototype explanations significantly improved counterfactual simulatability across both text and tabular data, indicating their unique effectiveness in helping users understand and predict model behavior when input data are perturbed (more on this below).

**2.3.5 Conflicting Results and the Import of Communicating Explanations**

How do we explain this result that feature graphs were shown by Hase and Bansal to be effective in enhancing deployer understanding (specifically, compared to Anchor explanations), when the studies above demonstrated that feature graphs were *ineffective* in increasing epistemic trust and participant accuracy? These contrasting results can be elucidated by considering three key differences between the related studies. First, participants in Hase and Bansal's study were undergraduate students with at least one course in computer science or statistics, which made them better equipped to utilize feature graphs effectively compared to participants in other studies who had no technical training.[12] Second, Hase and Bansal measured simulatability rather than accuracy, which might be more easily achieved with fluency in reading feature graphs than making accurate predictions about classifications on the basis of feature graphs. Finally, as these studies demonstrate, understanding the epistemic relevance of feature graphs incurs a cognitive load, the amount of which will vary depending on the participant's educational background (among other things). Hase and Bansal's participants, with their specific training, might have found the cognitive load manageable, whereas participants in other studies likely struggled.

With this in mind, let us explain the conflicting results between the studies by Hase and Bansal and Ribeiro and colleagues by examining the relative success that if-then statements had in Ribeiro's test of simulatability to the relative success that feature graphs had in Hase and Bansal's test of simulatability. In Hase and Bansal's study, participants were evaluated depending on whether they could accurately predict the model's behavior not only for novel inputs but also for *counterfactual scenarios*, wherein certain input features were perturbed. But anchor explanations are not well suited for counterfactual reasoning, as one cannot deduce facts about the consequent from the negation of the antecedent (the well-known 'fallacy of the inverse'). In other words, one cannot infer whether a change in input will necessarily lead to a different outcome simply by negating the input highlighted in the anchor rule.

---

[12] The exception to this is Weerts et al. (2019), which used undergraduates in computer science too. But the result that model confidence did not improve accuracy compared to feature graphs can be explained by poor calibration.

That said, Hase and Bansal also did not observe significant improvement in forward simulatability for Anchors, when inputs are novel rather than perturbed. This result is in direct conflict with how Ribeiro and colleagues found that Anchor explanations help deployers predict model behavior on novel inputs. But no elucidation is offered for this conflicting result unfortunately and seeing as the same method for generating Anchor explanations was applied in both studies, one can only point to differences in how Anchor explanations were *communicated*. Whereas Ribeiro and colleagues chose to display Anchor explanations with clear conditional language (e.g., labelling "if" and "then" clauses), Hase and Bansal chose a more technical manner of display, one which did not label the sides of the conditional, only opaquely rendering explanations as if-then rules to be used in natural deduction. Thus, again, the effectiveness of explanations amounts to their epistemic transparency, how well they lend themselves to ordinary epistemic practices.

### 2.3.6 Prototype Explanations and Counterfactual Simulatability

Let us now the effectiveness of prototype explanations in facilitating counterfactual simulatability. Prototype explanations enable deployers to compare new instances to representative examples from the training data. First, the model presents a prototypical example to describe the case under investigation (e.g., "Most Similar Prototype: Routine and rather silly"), then it provides a similarity score for how similar the current input is to the prototype (e.g., "Similarity score: 9.96 out of 10"). Crucially, these explanations are so effective because they facilitate ordinary epistemic practices. For example, a doctor might assess the likelihood of a diagnosis based on symptom similarities to known cases, while a lawyer might predict the court's decision by considering legal precedent. Moreover, if either person were to explain their decision-making, each could provide epistemic justification (all things being equal) by appealing to the process of comparing current cases to known examples. Thus, prototype explanations make model decisions transparent to deployers by aligning outputs with ordinary epistemic practices.

Thus, the reason why prototype explanations were more effective in facilitating counterfactual simulation than if-then rules is because they prompt abductive reasoning, drawing probabilistic inferences based on comprehensive understanding of several relevant cases. This is not to say that employing probabilistic reasoning safeguards either the doctor or the lawyer from the fallacy of the inverse. For example, the doctor might mistakenly believe that the probability of having disease X given symptom Y is the same as the probability of symptom Y given disease X. It is to say, however, that prototype explanations allow deployers to consider a wider range of possibilities (the prototype, the features making it prototypical, other cases that might fall under the prototype), which is essential for accurate counterfactual reasoning. Therefore, the success of different explanation methods can be attributed to the specific demands of the tasks and the type of reasoning required: if-then rules for forward simulatability and concrete examples for counterfactual simulatability.

**2.3.7 Addressing Cognitive Biases in Deployers of AI-DSS**

Before concluding our analysis of the literature, let us discuss a serious worry. Increasing epistemic trust in AI-DSS could lead to *blind* trust and overreliance, resulting in deployers refraining from engaging their decision-making processes, foregoing their own knowledge, and succumbing to cognitive biases. For example, Ghai et al. (2020) discovered that displaying model confidence produced an anchoring bias in participants with limited domain knowledge, predisposing them to endorse the model's predictions. Though this should not worry us too much, since deployers of AI-DSS are generally domain experts with extensive knowledge in their discipline, it does raise concerns about whether explanations can be "too effective".

The good news is that various strategies exist for mitigating cognitive biases when displaying explanations. For instance, Ma et al. (2023) demonstrated that overreliance on model confidence occurs only if it is displayed alone, without any accompanying reasoning for the prediction; meaning, overreliance was not observed if model confidence was shown alongside relative accuracy, or if displayed after participants made their predictions. These findings suggest that overreliance can be mitigated by pairing model confidence with additional explanations or withholding its display until after deployers have made their own predictions. So, we ought to be cautious about simply displaying confidence, and we need to consider carefully when to display it and what to accompany it with.

In one final study with a significant bearing on mitigating cognitive biases in deployers' use of AI-DSS, Chen et al. (2023) conducted a mix-methods study with IPT. By utilizing a think-aloud protocol to explore how participants utilized the two kinds of explanation, they found anchoring biases when deploying feature graphs, but not example-based explanations. Example-based explanations present profiles of two individuals with similar properties and predictions, thus functioning akin to prototypes, providing concrete examples that help deployers understand model behavior by comparing novel to known cases.

What is most fascinating about Chen and colleagues' study is that they found example-based explanations enabled participants to assess the accuracy of model predictions in a rigorous and reflective manner, and thus avoid anchoring biases associated with feature graphs. Analysis of the qualitative results demonstrated that participants explicitly reasoned through the examples, determining evidence and justification for their conclusions. Thus, to elucidate these results, Chen and colleagues appeal to the influential dual systems theory of cognition (Kahneman 2011), according to which cognitive biases result from exploits in the intuitive system (System 1) (e.g., feelings of fluency), exploits which can be avoided or overcome by the rational system (System 2) (e.g., natural deduction). Essentially, example-based explanations are able to avoid anchoring biases because they facilitate sound epistemic practices.

In conclusion, the studies above reveal the importance of designing AI-DSS to serve as epistemic quasi-partners in decision-making, whose contributions facilitate natural inferential processes and promote analytical reasoning. Clear, declarative explanations in propositional formats and rule-based conditionals are effective in enhancing trust calibration and participant accuracy by aligning outputs with ordinary epistemic practices. By promoting analytical reasoning and engaging the rational cognitive system, example-based, counterfactual descriptions, and prototype explanations reduce the risk of depending on the intuitive, bias-prone cognitive system. Thus, AI-DSS ought to be designed to enhance human reasoning processes by offering explanations that can be fluently employed in ordinary epistemic practices, a point we return to below (**Section 4**). For now, let us turn to current XAI theories of good human-grounded explanations.

## 3. The Inadequacy of Current XAI Theories

In this section, we canvass three prominent theories that bear on what makes for good human-grounded reasons. All theories (including our own) have a guiding principle in common: *the quality of a human-grounded explanation can and ought to be determined by measuring its effect on trust calibration*. To make the argument more concise, a gloss is applied to the above principle, focusing scrutiny on one side of the proverbial equation, namely on perceived trustworthiness. Thus, the semi-principle reads, *the quality of a human-grounded reason can and ought to be determined by measuring its effect on perceived trustworthiness*. These principles are only meant to be employed in the domain of AI-DSS, and, as such, they assume that the provider of the reason (the machine) is incapable of lying or intentionally misleading. We argue that current theories either do not adequately explain the evidence above or do not impart sound ethical advice for development.

## 3.1 Theory of Mental Models

**Theory of Mental Models**

Premise 1: Developing trust in AI-DSS requires developing a mental model.

Premise 2: Good mental models consist of explanations of how AI-DSS generate outputs.

Conclusion: AI-DSS will be perceived as trustworthy if deployers are given explanations for how the system generates outputs.

The theory of mental models (TMM) posits that perceived trust in AI-DSS can be increased through explanations for how the model works (e.g., Hoffman et al. 2018a, Hoffman et al. 2018b). That said, "how it works" is too vague to be of use in philosophical analysis, so let us place the following gloss on it, namely how it *generates outputs*. Albeit an interpretation, it is nonetheless faithful to TMM, since the theory calls for evaluating competing explanations based on how well they facilitate deployers predicting model outputs. Though to our knowledge there is no philosophical defense of TMM, one can readily intuit its line of reasoning. It infers its conclusions from two premises: first, developing trust in AI-DSS necessitates the construction of mental models; second, effective mental models are constructed from explanations of how AI-DSS generate outputs.

One serious issue with this manner of spelling out what constitutes a good human-grounded explanation is that it is unclear whether it offers a meaningful difference between explainability and interpretability. This is further underscored by how TMM focuses on mechanistic explanations that aim to elucidate the internal processes and decision-making pathways (Hoffmann et al. 2018b), which may not be useful or easily understandable for typical deployers. In other words, by concentrating on explanations of how models generate outputs (how they work), TMM falls into the trap of maintaining that explanations that are good for developers are also good for deployers, which runs afoul of the lunatics-running-the-asylum problem and renders human-grounded explanations incoherent.

Regarding its adequacy in explaining the evidence above, TMM lacks the conceptual tools for elucidating the remarkable effectiveness that displaying model confidence and counterfactual and prototypical descriptions exert in increasing perceived trustworthiness. TMM maintains that explanations ought to reveal the mechanics of the model, how certain mechanisms generate certain outputs, but model confidence and the like do not reveal how the model yielded its prediction — they make the results of relevant mechanisms comprehensible by aligning them with how humans explain their own decision-making. Other issues confront TMM as well. It is counterintuitive: deployers routinely trust automation whose inner workings remain opaque to them. It is conceptually vague: it is unclear about the sufficiency of detail within the mental model. Finally, it is impractical, suggesting deployers acquire the expertise of developers.

## 3.2 Theory of Mental Models Plus

**Theory of Mental Models Plus**

Premise 1: Developing trust in AI-DSS requires building a mental model.

Premise 2: Good mental models consist of explanations about when AI-DSS makes errors.

Conclusion: AI-DSS will be perceived as trustworthy if deployers are given explanations about their error boundaries.

The next theory of good human-grounded explanations builds upon TMM, offering an augmented version that places particular emphasis on error boundaries. Bansal and colleagues (2019) offer a formal definition of an error boundary: "The error boundary of model $h$ is a function $f$ that describes for each input $x$ whether model output $h(x)$ is the correct action for that input: $f: (x, h(x)) \rightarrow \{T, F\}$" (3). Thus, an error boundary tells you which predictions made by the model are correct and which are incorrect, like a map that marks the places where the model yields true outputs and false outputs, respectively. Call this the Theory of Mental Models Plus (TMM+).

Similar to TMM above, we can understand TMM+ as positing two fundamental premises: first, developing trust in AI-DSS necessitates the construction of a mental model; second, good mental models are comprised of explanations detailing circumstances under which AI-DSS make mistaken outputs. From these premises, it follows that AI-DSS will be perceived as trustworthy if deployers are provided with explanations regarding their error boundaries, offering insight into when and how AI-DSS make mistaken outputs.

TMM+ faces similar criticism as TMM for being counterintuitive and conceptually limited. Deployers often trust technology without understanding its error boundaries. To use a simplified example, you rely on your automobile to bring you to work, but you likely do not know the technical condition under which the engine would fail to start. TMM+ is also conceptually limited: because it focuses on errors in outputs, TMM+ does not have the conceptual resources to explain why, for example, errors in confidence have the potential to be so informative. Intuitively, knowing when the model is under or overconfident reflects an important error boundary, and this kind of error is distinct from what is captured by the above definition.

Regarding the evidence above, TMM+ also struggles to explain the effectiveness of model confidence, counterfactual and prototype descriptions. This is because simply knowing the confidence behind a prediction does not amount to knowledge of how the model produces mistaken outputs. Confidence scores only have a bearing on developing error boundaries (in a more general sense than how Bansal and colleagues defined it) if deployers also detect under or overconfidence. But even determining errors in confidence calibration will not inform about the underlying causes of miscalibration, which is required for the development of general boundaries of when the model errs. Thus, while necessary, model confidence is not sufficient for the development of accurate error boundaries of the model, meaning TMM+ lacks the conceptual tools to explain why model confidence is so effective.

Another criticism concerns TMM+ lack of conceptual resources to explain the effectiveness of counterfactual and prototypical explanations. While counterfactuals offer insight into error boundaries, since they reveal information about thresholds around decision, they may not accurately pinpoint instances where the model errs when changing its decision. For example, a counterfactual explanation might show that changing an input, such as a loan applicant's income from $50,000 to $60,000, results in a loan approval. However, it does not elucidate why an applicant with an income of $55,000, who is otherwise similar in all other aspects, was denied. Deployers are left to speculate about the cause, whether it reflects inconsistencies or biases in the model, that is, whether this output, coupled with this counterfactual explanation, are evidence of a mistaken output. We also need to know the internal decision-making process, including feature importance and the contribution of latent variables, in order to determine whether the model has indeed erred. Finally, since prototype-based explanations simply display a previous example, the classification, and a similarity score, it does not offer any information that would assist in determining the model's error boundaries. Consequently, TMM+ lacks the conceptual tools to explain why these explanations are so effective in increasing perceived trustworthiness.

## 3.3 Theory of Anthropomorphism

**Theory of Anthropomorphism**

Premise 1: Human behavior is trusted if it is explainable (i.e., rational, reasonable, justified).

| | | |
|---|---|---|
| Premise 2: | | People anthropomorphize AI-DSS, treating them as though they were human. |
| Conclusion: | | AI-DSS will be perceived as trustworthy if deployers anthropomorphize AI-DSS as human-like decision-makers, providing rational, reasonable, and justified explanation for their behavior. |

Another prominent theory that elucidates the relationship between explanations and perceived trustworthiness in AI-DSS is the theory of anthropomorphism (TA) (e.g., Seymour and Kleek 2021). TA can be seen as positing two fundamental premises: first, human behavior garners trust if it is explainable, meaning rational and reasonable; second, individuals anthropomorphize AI-DSS, attributing to them human-like characteristics. From these premises, it follows that AI-DSS will be perceived as trustworthy if AI-DSS are anthropomorphized to behave like human agents wherever tasked with explaining their decision-making.

The advantage that TA has over TMM and TMM+ is that it has the conceptual tools to explain the evidence discussed above. Indeed, humans explain their decision-making by offering propositional descriptions constrained by truth conditions. We also tend to describe our confidence levels while making decisions under uncertainty, and if we are placed in a position to justify our decisions, we tend to offer examples and counterexamples. Each one of these explanatory features is predicted by TA and confirmed by the empirical evidence on XAI.

But is it ethical to design machines to be anthropomorphic? The success of this design strategy could be attributable to exploiting cognitive biases in human decision-making, namely in-group biases, trusting agents because they look and talk like us. Another way of understanding what is problematic about this strategy is that it might amount to instilling a category mistake in the mind of deployers (a fallacy in reasoning), causing them to ascribe machines human-like qualities that are beyond, or even alien to, their actual capacities. In other words, designing anthropomorphic technology for the purpose of increasing perceived trustworthiness, without any independent reasons for doing so, can be seen as manipulating trust perception rather than facilitating it.

Consequently, a novel theory of human-ground explanations must possess the explanatory dexterity to navigate between Scylla and Charybdis: it must avoid being ensnared by the six-headed beast, whose many anthropomorphic heads chew theories up for their unethical council, while also steering clear of the whirlpool, which drags theories down into its depths of equivocations between interpretability and explainability. With this task in mind, let us now consider the theory of epistemic quasi-partnerships.

## 4. The Theory of Epistemic Quasi-Partnerships and the RCC Approach

**Theory of Epistemic Quasi-Partnerships**

Premise 1:  AI-DSS are epistemic quasi-partners, offering unique and expert contributions to a decision space without the social responsibility of genuine partners.

Premise 2:  Epistemic partners (whether quasi or genuine) are perceived as epistemically trustworthy if they engage in sound epistemic practices when explaining their decision-making.

Conclusion:  AI-DSS will be perceived as epistemically trustworthy if they are designed to engage in sound epistemic practices.[13]

In this section, we motivate EQP by appealing to a thought experiment and providing an initial defense of the theory as adequately describing the empirical evidence above. Let us begin by addressing what it means to call AI-DSS *quasi*-partners. As briefly remarked above, quasi means 'as though', and so we wish to convey that the status of AI-DSS as epistemic partners is a kind of *fiction*, one justified by its explanatorial efficacy. This is because genuine partners are socially responsible and AI-DSS are not the kind of entity that can be held socially responsible (see Dorsch and Deroy 2024 for an extensive discussion of this). In reality, AI-DSS are *tools for decision support* and their unique capacity to contribute valuable information to the decision space lends itself to thinking of them *as though* they were epistemic partners, a particularly advantageous fiction for determining how best to explain their role in hybrid decision-making environments. Though a full defense of this position would go beyond the present scope, the thought experiment below should motivate adopting a fictional stance toward AI-DSS.

Call this thought experiment: *Crossroads at the AI-Assisted Trail*. You and your partner are hiking through tricky mountainous terrain when you arrive at a crossroads and need to determine the best path forward. Your partner is a technical expert skilled in utilizing navigation technology with no knowledge of the terrain, and you are a domain expert with extensive experience in navigating this landscape with no technical expertise. You represent the deployer, who relies on firsthand knowledge and intuition, while your partner represents the AI-DSS, employing computational models and issuing predictions. Now, imagine you two disagree about the best path forward. What do you need to hear from your partner before you can change your mind?

---

[13] Some examples of sound epistemic practices are giving and getting reasons, offering and responding to counterexamples, and eliciting and describing confidence levels.

As the domain expert, three factors are essential for switching to the technical expert's prediction. First, you need access to the domain-relevant reasons behind their decision-making. This transparency enables you to determine whether their prediction is justified and warranted given the evidence. Secondly, you need to engage in rational dialogue with them, wherein you can determine how their prediction changes in response to counterexamples; such a dialogue ensures various contingencies are considered, contributing to the robustness of the collective decision. Finally, knowledge of their confidence levels will provide an insight into the reliability of both their predictions and predictive processes. Hence, each factor reflects a hypothesis regarding the underlying causes of perceived trustworthiness of AI-DSS: reasons, counterfactuals, and confidence (RCC).

In advocating for the EQP, several key components contribute to the perception of AI-DSS as trustworthy entities within collective decisions. The first premise posits that AI-DSS serve as epistemic quasi-partners, offering unique and expert contributions to the decision-making process without the social responsibility of genuine partners. The second premise highlights the essential characteristics of epistemically trustworthy partners, whether genuine or otherwise: engaging in sound epistemic practices around explaining their decision-making. These characteristics are encapsulated in the RCC approach, which is a non-exhaustive outline containing three predictions regarding when model behavior will be perceived as epistemically trustworthy by deployers. According to this approach, epistemic partners are perceived as trustworthy if they provide reasons behind their decisions (these can come in the form of example-based reasons), demonstrate responsiveness to and appropriate alignment with counterfactuals, and maintain transparency regarding confidence levels.

**RCC Approach**

| | |
|---|---|
| Reasons: | Know the reasons behind the technical expert's decision. <br> "I made this decision because…" |
| Rational Dialogue (sensitive to *Counterfactuals*): | How the decision changes wrt counterfactuals: <br> "If things were different, what would this mean for the decision?" |
| Confidence: | Know their confidence, preferably wrt counterfactual changes. <br> "I am 90% confident, but if things were different, I would only be 40%." |

For example, "The recommendation is to perform $\varphi$ (confidence at x) because P, but if it were not the case that P, then confidence in $\varphi$ would drop to y, and the recommendation would be to perform $\theta$ (confidence at z)", whereby x, y and z are whole-number percentage values between 0 and 100 that reflect the degree of confidence that the system has in its decision-making (e.g., its precision, recall or accuracy score). Two limitations to both EQP and the RCC approach, however, are that they apply only when the system's predictions are explainable in human terms, and it is possible to provide reliable confidence scores. Meanwhile, a recent ally of RCC is Milani and colleagues' (2023) Bayesian network approach, which aims to enhance the perceived trustworthiness of AI by providing clear reasons behind decisions.

Does EQP adequately explain the empirical evidence above? Similar to how TA explains the evidence by appealing to properties of human decision-makers, EQP appeals to properties of sound epistemic practices common to human culture. Does EQP offer unethical advice centered around anthropomorphizing machines? Unlike TA, EQP stresses that AI-DSS are *quasi*-partners, who lack the social responsibility of genuine partners, and so EQP has levers for prying conceptual distance between human-human and human-AI interactions. That said, it has not been demonstrated whether EQP has the conceptual liquidity to pay the debt incurred by increasing perceived trustworthiness. Could the aim of increasing trustworthiness in machines (arguably a trait that can be only had by our peers (Bryson 2018) or morally rational agents (Dorsch & Deroy 2024)) be itself unethical? We leave addressing this concern for another time.

## 5. Conclusion

The theory of epistemic quasi-partnerships offers a compelling framework for understanding epistemic trust in AI-DSS, as it aligns with empirical evidence, is conceptually clear, and provides practical and ethical guidance. By emphasizing the importance of epistemic quasi-partnerships and the RCC Approach—where reasons, counterfactuals, and model confidence are prioritized—the theory provides a comprehensive explanation for how epistemic trust is cultivated in hybrid decision-making. Thus, the RCC approach is not only an effective means of improving trust calibration in accurate systems, demonstrated by the wealth of empirical support discussed above, it is also an ethical approach to improving epistemic trust in machine-based systems, since it does not advise developing anthropomorphic machines and could lead to mitigating algorithm aversion.

## Authors' Contributions

The manuscript was written by John Dorsch. Maximilian Moll provided critical analysis on the manuscript, as well as commentary, technical and practical perspective on machine learning methods. All authors read and approved the final manuscript.

## Acknowledgments

This work was supported by Bayerisches Forschungsinstitut für Digitale Transformation (bidt) Co-Learn Award Number KON-21-0000048.